%% file: egpaper_for_review.tex
\documentclass[10pt,twocolumn,letterpaper]{article}

\usepackage{iccv}
\usepackage{times}
\usepackage{epsfig}
\usepackage{graphicx}
\usepackage{amsmath}
\usepackage{amssymb}

\usepackage{diagbox}
\usepackage{array}
\usepackage{booktabs}
\usepackage{multirow}
\usepackage{booktabs}
\usepackage{bbding}
\usepackage{pifont}
\usepackage{tipa}
\usepackage{amsmath}
\usepackage{xcolor}
\usepackage[table]{xcolor}
\usepackage{enumitem}
\usepackage{wrapfig}
\usepackage[accsupp]{axessibility}  

\definecolor{cvprblue}{rgb}{0.21,0.49,0.74}
\usepackage[pagebackref,breaklinks,colorlinks,allcolors=cvprblue]{hyperref}

\iccvfinalcopy 

\def\iccvPaperID{2811} 

\definecolor{Gray}{gray}{0.5}
\definecolor{LightCyan}{rgb}{0.88,1,1}

\newcolumntype{a}{>{\columncolor{Gray}}c}
\newcolumntype{b}{>{\columncolor{white}}c}

\ificcvfinal\fi

\begin{document}

\title{Seeing 3D Through 2D Lenses: 3D Few-Shot Class-Incremental Learning via Cross-Modal Geometric Rectification}
\vspace{-6mm}
\author{%
Tuo Xiang$^{1}$\quad Xuemiao Xu$^{1,2,3,4}$\quad Bangzhen Liu$^{1}$\footnotemark[1]\quad Jinyi Li$^{1}$\quad Yong Li$^{1}$\footnotemark[1]\quad Shengfeng He$^{5}$ \\
\normalsize$^{1}$South China University of Technology \quad$^{2}$State Key Laboratory of Subtropical Building Science\\
\normalsize$^{3}$Guangdong Provincial Key Lab of Computational Intelligence and Cyberspace Information\\
\normalsize$^{4}$Ministry of Education Key Laboratory of Big Data and Intelligent Robot \quad\normalsize$^{5}$Singapore Management University\\
}
\maketitle
\ificcvfinal\thispagestyle{empty}\fi

\renewcommand{\thefootnote}{\fnsymbol{footnote}}
\footnotetext[1]{Corresponding authors (liubz.scut@gmail.com, yong@scut.edu.cn).}

\begin{abstract}
The rapid growth of 3D digital content necessitates expandable recognition systems for open-world scenarios. However, existing 3D class-incremental learning methods struggle under extreme data scarcity due to geometric misalignment and texture bias. While recent approaches integrate 3D data with 2D foundation models (e.g., CLIP), they suffer from semantic blurring caused by texture-biased projections and indiscriminate fusion of geometric-textural cues, leading to unstable decision prototypes and catastrophic forgetting.
To address these issues, we propose Cross-Modal Geometric Rectification (CMGR), a framework that enhances 3D geometric fidelity by leveraging CLIP’s hierarchical spatial semantics. Specifically, we introduce a Structure-Aware Geometric Rectification module that hierarchically aligns 3D part structures with CLIP’s intermediate spatial priors through attention-driven geometric fusion. Additionally, a Texture Amplification Module synthesizes minimal yet discriminative textures to suppress noise and reinforce cross-modal consistency. To further stabilize incremental prototypes, we employ a Base-Novel Discriminator that isolates geometric variations.
Extensive experiments demonstrate that our method significantly improves 3D few-shot class-incremental learning, achieving superior geometric coherence and robustness to texture bias across cross-domain and within-domain settings. 
\end{abstract}

\input{sections/1_Introduction}
\input{sections/2_RelatedWork}

\input{sections/3_Method}
\input{sections/4_Experiments}
\input{sections/5_Conclusion}

{\small
\bibliographystyle{ieee_fullname}
\bibliography{egbib}
}

\end{document}

%% file: sections/1_Introduction.tex
\section{Introduction}

\begin{figure}
\centering
\includegraphics[width=1\columnwidth]{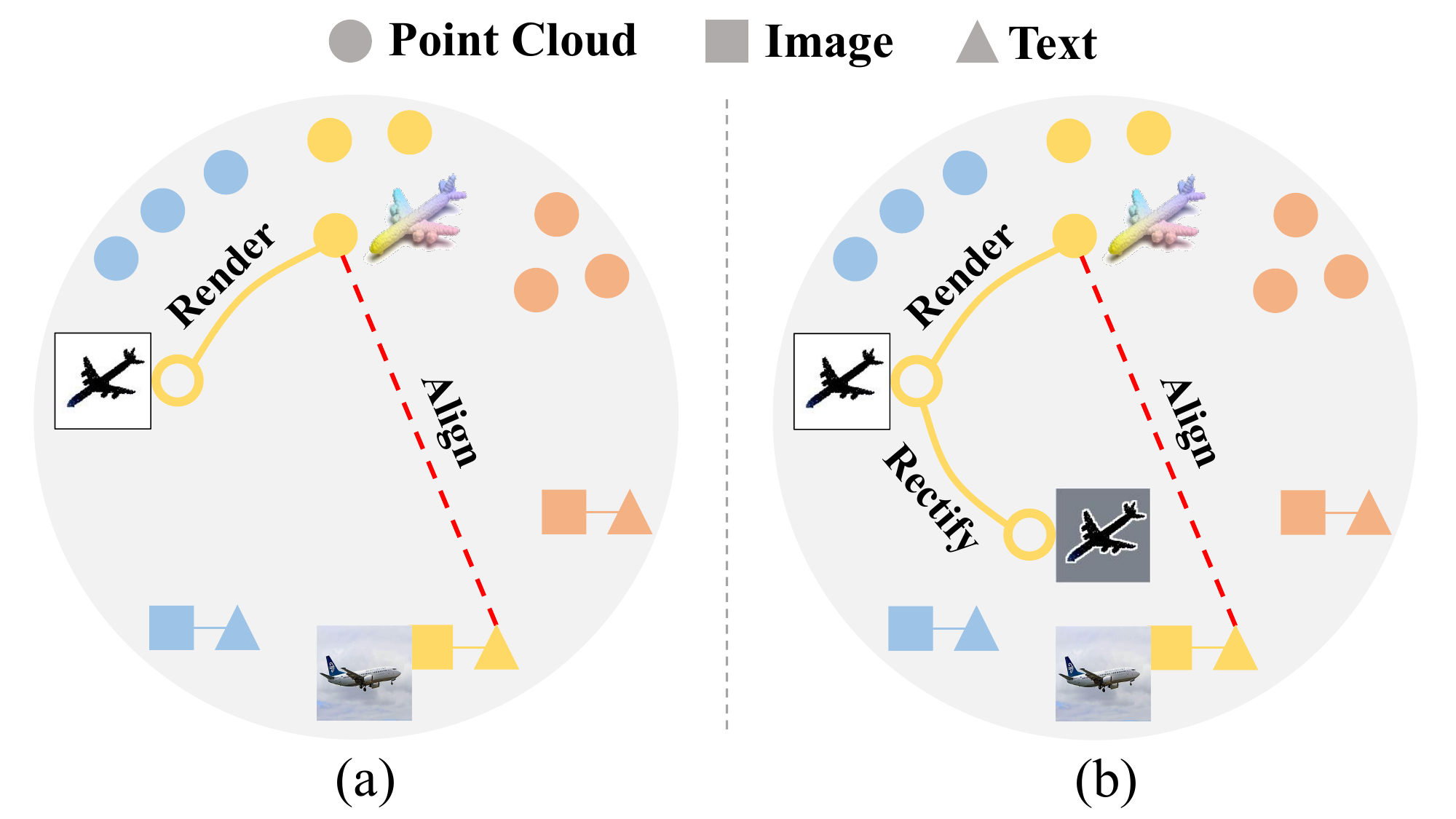}
\caption{
Direct alignment of point clouds with CLIP’s text features proves challenging due to the inherent domain gap. (a) While projecting depth maps to leverage 2D priors can bring point cloud and text features closer, discrepancies between the depth map and CLIP’s input space impede effective alignment. (b) In contrast, through 2D-3D rectification, point clouds are enriched with hierarchical 2D information derived from CLIP, facilitating a more precise alignment with text features.}
\label{fig:align}
\end{figure}

The proliferation of 3D digital content~\cite{chang2015shapenet, wu20153d, reizenstein2021common, uy2019revisiting, deitke2023objaverse, zheng2025recdreamer, liu2025genpoly, xu2024learning, lu20243d} has revolutionized environmental modeling in domains such as autonomous navigation and industrial robotics. However, existing 3D recognition systems~\cite{qi2017pointnet, qi2017pointnet++, wang2019dynamic, liu2025rotation} remain ill-equipped for open-world deployment, where novel object categories may dynamically emerge under extreme data scarcity, often leading to catastrophic degradation of performance. This limitation stems from a fundamental mismatch between static latent representations and the evolving geometric complexity of real-world 3D environments. Addressing this challenge necessitates progress in 3D Few-Shot Class-Incremental Learning (3DFSCIL), a setting in which models must incrementally integrate sparse geometric cues from limited point cloud data while preserving discriminative capabilities for previously encountered categories.

Earlier 3DFSCIL approaches~\cite{chowdhury2022few} employed low-dimensional embeddings derived from PCA or SVD to mitigate catastrophic forgetting. While effective in reducing overfitting on scarce incremental samples, they did so at the expense of geometric expressiveness, particularly in the context of structurally complex point clouds. More recent strategies~\cite{xu2023filp, cheraghian2024canonical} have leveraged 2D vision foundation models such as CLIP~\cite{radford2021learning} by incorporating depth map rendering~\cite{xu2023filp} or synthetic texture augmentation~\cite{cheraghian2024canonical} to bridge modality gaps. However, these methods depend on single-step 3D-to-2D projections that extract only CLIP’s final-layer representations, discarding the rich spatial priors encoded in intermediate layers—priors that more naturally correspond to structural 3D hierarchies. This omission introduces semantic misalignment between novel 3D categories and their 2D projections (Figure~\ref{fig:align}(a)), exacerbating semantic blurring due to the naive multimodal fusion mechanism.
We posit that robust cross-modal alignment in 3DFSCIL hinges on a crucial yet underutilized insight: CLIP's intermediate layers preserve spatial relationships (e.g., part adjacency and symmetry) that are inherently aligned with 3D geometric hierarchies. However, the exclusive focus on CLIP’s final layer of existing methods inadvertently forces depth maps into texture-biased semantics space and erodes geometric distinctiveness. Furthermore, conventional adapters indiscriminately fuse visual and geometric features, collapsing 3D representations into texture-dominated spaces. This dual misalignment induces geometric myopia, wherein incremental updates amplify texture-driven biases while suppressing structural coherence.

To address these challenges, we propose cross-modal geometric rectification, which repurposes CLIP’s intermediate layers as ``2D lenses'' to enhance geometric fidelity. Our framework corrects three core misalignments. First, it mitigates semantic blurring by integrating 3D point cloud features with CLIP’s hierarchical spatial priors rather than relying on texture-biased final-layer representations. Second, it resolves modality mismatch through semantic-aware texture synthesis, aligning depth maps with CLIP’s pretrained space without corrupting structural integrity. Third, it stabilizes incremental learning by dynamically reconfiguring decision boundaries to isolate geometric variations from historical knowledge.
Specifically, our approach consists of three key components: the Structure-Aware Geometric Rectification (SAGR) module, the Texture Amplification Module (TAM), and the Base-Novel Discriminator (BND). The SAGR module hierarchically integrates CLIP’s intermediate features with 3D structural embeddings through attention mechanisms that prioritize part-level coherence over texture cues. Meanwhile, the TAM learns minimal yet discriminative texture patterns to align depth maps with CLIP’s visual space, eliminating ambiguous texture extraction in geometric rectification. Finally, the BND isolates geometrically distinct regions for novel classes, ensuring stable knowledge retention across incremental phases. By transforming CLIP from a texture-driven projector into a geometry-refining lens, our framework sharpens 3D representations, enabling precise cross-modal alignment and improved incremental learning under extreme data scarcity.

In summary, our contributions are threefold:
\begin{itemize}
    \item We propose Cross-Modal Geometric Rectification, a framework that leverages CLIP’s hierarchical 2D semantics to enhance 3D representations through structural fusion of intermediate spatial priors and 3D geometric embeddings.
    \item We introduce a Structure-Aware Geometric Rectification module that hierarchically aligns 3D part structures with CLIP’s intermediate spatial priors via attention-driven geometric fusion, along with a Texture Amplification Module that synthesizes minimal yet discriminative textures to reinforce cross-modal consistency while suppressing noise.
    \item Extensive experiments validate our method’s effectiveness in mitigating texture bias and preserving geometric coherence, achieving state-of-the-art performance in 3D few-shot class-incremental learning across both cross-domain and within-domain settings.
\end{itemize}

%% file: sections/2_RelatedWork.tex
\section{Related Work}

\subsection{Few-Shot Class Incremental Learning}
Early FSCIL methods focused on 2D images, graph-based classifier adaptation~\cite{tao2020few} was first proposed and followed by replay-based~\cite{kukleva2021generalized,liu2022few,agarwal2022semantics, xu2025fr2seg} and pseudo-based~\cite{zhou2022forward,peng2022few,zhang2021few,zhu2021self} methods. Attention mechanisms were further used in~\cite{zhang2021few,zhou2022few,chi2022metafscil,zhao2023few}. Subsequent 3D extensions like I3DOL~\cite{dong2021i3dol} introduced geometric centroid alignment for point cloud incremental learning, while MicroShape~\cite{chowdhury2022few} leveraged text embeddings to enhance few-shot feature transfer. Recent advances exploit vision-language models: FILP-3D~\cite{xu2023filp} pioneers CLIP integration via depth map rendering, and C3PR~\cite{cheraghian2024canonical} improves CLIP adaptation through viewpoint reprogramming. However, these methods rely on single-step 3D-to-2D projections that over-prioritize CLIP's texture-biased final layers, neglecting intermediate spatial priors critical for geometric hierarchy alignment. This oversight perpetuates semantic ambiguity between modalities, particularly in disentangling structural coherence from texture noise during incremental updates.

\subsection{Foundation Models}
Foundation models have redefined artificial intelligence through extensive pretraining on diverse datasets. BERT~\cite{devlin2019bert} has advanced natural language processing via large-scale masked language modeling, while GPT-3~\cite{mann2020language} has demonstrated notable few-shot learning in language generation. In computer vision, CLIP~\cite{radford2021learning} integrates visual and textual modalities through training on expansive image-text pairs, thereby driving progress in multimodal learning. More recently, Uni3D~\cite{zhou2023uni3d} introduced a unified, scalable framework for 3D pretraining, enabling large-scale 3D representation learning and extending the reach of foundation models to 3D tasks. These developments highlight the promise of foundation models for robust generalization across varied domains, though their application to 3DFSCIL tasks remains in its early stages.

\begin{figure*}
\centering
\includegraphics[width=2\columnwidth]{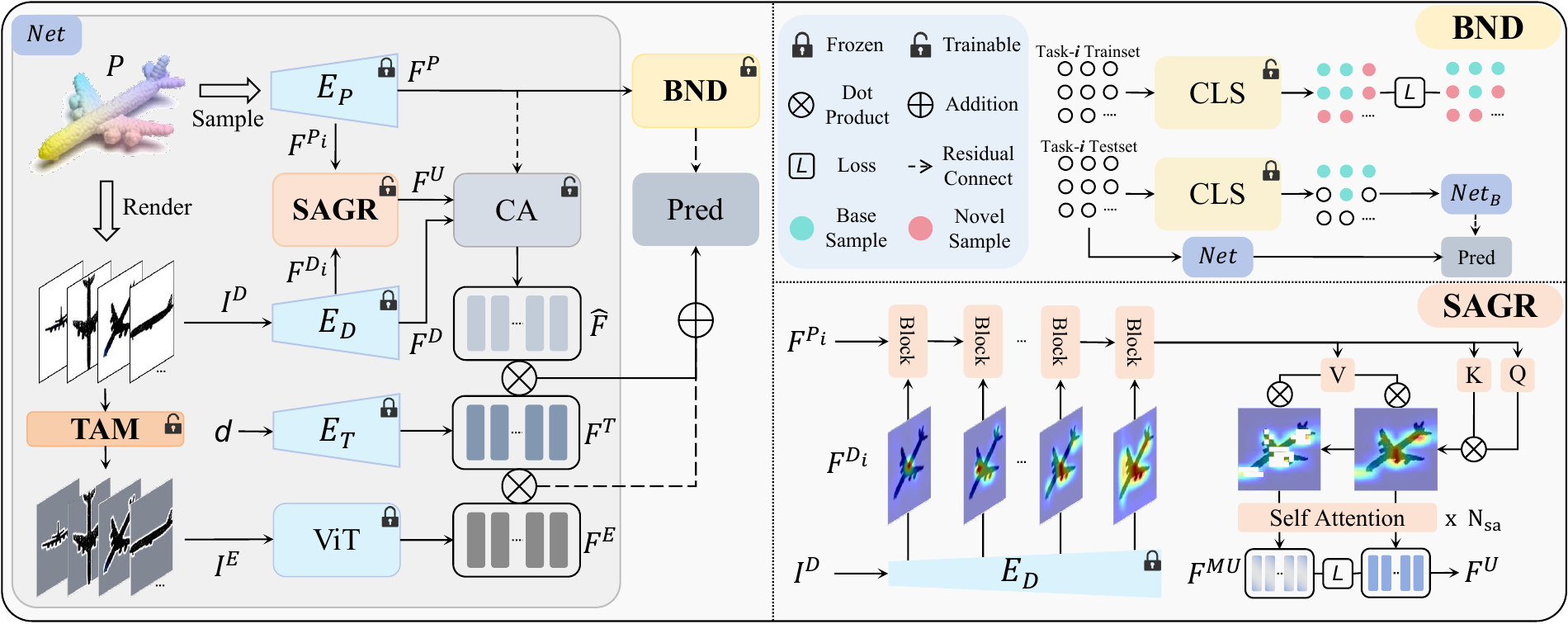}
\caption{Overview of our proposed framework. CMGR reveals a two-stream processing approach for point clouds, comprising a 3D branch and a 2D branch. In the 3D branch, the point cloud is first processed by a transformer-based point encoder $E_P$ to generate intermediate and final feature representations. The intermediate features $F^{P_i}$ are subsequently aligned with the depth encoder $E_D$ through the Structure-Aware Geometric Rectification module, while the final features $F^P$ are passed to subsequent modules for further processing. In the 2D branch, the depth maps $I^D$ are projected, processed by $E_D$, and then enhanced via the Texture Amplification Module to refine predictions. For each incremental task, the Base-Novel Discriminator performs pre-classification, distinguishing base from novel classes, thereby enabling parallel processing of these categories.}
\label{fig:pipeline}
\end{figure*}

\subsection{Cross-Modal Learning for Point Clouds}
Cross-modal learning seeks to integrate diverse modalities to yield more robust representations. Pretrained models have become instrumental in extracting primary features from each modality. CLIP~\cite{radford2021learning} established a seminal approach by mapping image-text pairs into a shared feature space via contrastive learning. Building on this paradigm, CrossPoint~\cite{afham2022crosspoint} applies self-supervised contrastive learning to images rendered from point clouds, thereby enhancing the training of point cloud feature extractors for tasks such as classification and segmentation. Similarly, PointCLIP~\cite{zhang2022pointclip} and its successor, PointCLIP V2~\cite{zhu2023pointclip}, align point clouds with textual descriptions by processing rendered depth maps through the CLIP framework. CLIP2Point~\cite{huang2023clip2point} further improves the extraction of point cloud features by pretraining the CLIP Vision Transformer on depth maps, while subsequent studies~\cite{xue2023ulip,qi2023contrast} have advanced 3D understanding by integrating 3D point clouds, 2D images, and text. Notably, these methods generally employ a single encoder guided by a frozen pretrained model, which overlooks potential inter-modal interactions and thus limits overall representational capacity.

%% file: sections/3_Method.tex
\section{Method}

\subsection{Preliminaries}
Let $D^0=\{(x_j^0,y_j^0)\}_{j=1}^{|D^0|}$ denote the base point cloud dataset, comprising $|D^0|$ samples spanning $L_0$ distinct categories. Our goal is to learn a model $f_\theta$ from $D^0$ and subsequently adapt it to a sequence of incremental tasks $\mathbf{D}=\{D^1, D^2, ..., D^{T-1}\}$. Each task is defined as $D^t=\{(x_j^t,y_j^t)\}_{j=1}^{|D^t|}$ for $t\in \{1,..., T-1\}$ and contains only a few available samples from a novel set of categories $C_t$, such that $|D^0|\gg |D^t|$. Moreover, the label spaces are mutually exclusive, \ie, $C^i \cap C^j=\emptyset$ for any $i\neq j$ in $\{0, ..., T-1\}$. Following initial training on $D^0$, the model is incrementally updated by sequentially incorporating each task, with evaluation performed on a unified test set that contains all categories encountered up to the current task $t$.
\subsection{Overview}

To mitigate catastrophic forgetting and overfitting in 3DFSCIL, we introduce Cross-Modal Geometric Rectification (CMGR), a novel framework designed to improve the alignment of 3D point clouds with their corresponding prototypes. By employing hierarchical cross-attention between 2D and 3D features, CMGR enables a more refined integration of the hierarchical semantic information encoded in CLIP, thereby enhancing the geometric fidelity of the 3D representations. Furthermore, CMGR addresses the challenge of domain shift in CLIP’s input data through texture amplification, which refines depth map representations.

As shown in Figure~\ref{fig:pipeline}, CMGR consists of three modules: the Structure-Aware Geometric Rectification (SAGR) module, which integrates multi-level 3D and 2D features through cross-attention to refine 3D representations; the Texture Amplification Module (TAM), which synthesizes discriminative textures from 3D features to optimize CLIP’s predictions; and the Base-Novel Discriminator (BND), which stabilizes the incremental learning process by maximizing the classification boundary between base and novel classes, thus preserving geometric independence. By refining the fusion of geometric and textural information through CMGR, we ensure that both structural integrity and semantic coherence are preserved throughout the incremental learning process, leading to more stable and accurate performance in both cross-domain and within-domain settings.




\subsection{Structure-Aware Geometric Rectification} 
Traditional 3DFSCIL methods suffer from geometric misalignment due to their reliance on CLIP's texture-biased final-layer projections. To resolve this, our SAGR module establishes the hierarchical correspondence between 3D structural hierarchies and CLIP's intermediate spatial semantics. Unlike prior works that naively fuse final-layer 2D-3D features, we implement multi-stage geometric rectification through CLIP's layer-wise spatial priors.
The core innovation lies in our redesigned cross-modal attention mechanism. Given CLIP-encoded text prototypes $F^T$ from class descriptions $d$ and the extracted 2D features $F^P$/$F^D$ for point clouds/depth maps from the 3D/2D encoders $E_P$/$E_D$ that contain totally $N_l$ layers, we enforce geometric consistency through layer-adaptive feature fusion. For each transformer block $i \in \{0,\dots, N_l-1\}$:
\begin{align}
\text{Attn}(Q,K,V) = \text{softmax}\left(\frac{QK^\top}{\sqrt{d_k}}\right)V, \\
F^{P_{i+1}} = 
\begin{cases} 
\text{Attn}(F^{P_i}, F^{D_i}, F^{D_i}) & i \in L_r\\
\text{Attn}(F^{P_i}, F^{P_i}, F^{P_i}) & \text{otherwise}
\end{cases},
\end{align}
where $L_r$ denotes the selected CLIP layers that contain part-level spatial relationships. This selective fusion directly addresses the semantic blurring problem by preserving geometric partonomy through mid-level feature interactions. 

To enhance the learned geometric representation in cross-modal fusion, we implement self-masking attention with mask ratio $M_R$:
\begin{equation}
R^M[q,k] = 
\begin{cases} 
R[q,k] & R[q,k] < \text{Top}(M_R) \\
0 & \text{otherwise}
\end{cases}.
\end{equation}
$R[q, k]$ denotes an attention weight matrix with $q$ rows (query) and $k$ columns (key). Top($M_R$) refers to the element in $R$ whose attention weight ranks at the top $M_R$ percentile in descending order. We obtained masked features $F^{MU}\!=\!R^M\cdot V$ that explicitly suppress the geometric non-relevant attention patterns. Incorporating with the unmasked $F^U\!=\!R\cdot V$ features, we employ a regularization within $N_{sa}$ self-attention layers, which is denoted by:
\begin{equation}
\mathcal{L}_{mc} = \frac{1}{B^2}\sum_{i,j=1}^B \vert\vert\mathit{sim}(F_i^U,F_j^U) - \mathit{sim}(F_i^{MU},F_j^{MU})\vert\vert_2^2
\end{equation}
Finally, we incorporate the 2D-geometric rectified feature $F^U$ with the original features $F^D$ and $F^P$ via a cross-view aggregation operation (CA), which is formulated as follows:
\begin{equation}
\widehat{F} = f\left( (f'(\text{concat}(F^P,F^U)) + w \cdot F^D) \odot \lambda \right),
\end{equation}
where $w$ and $\lambda$ are weight factors to control the rectification ratio, $f$ and $f'$ represent the linear fusion function.

\subsection{Texture Amplification Module}

\begin{figure}
\centering
\includegraphics[width=1\columnwidth]{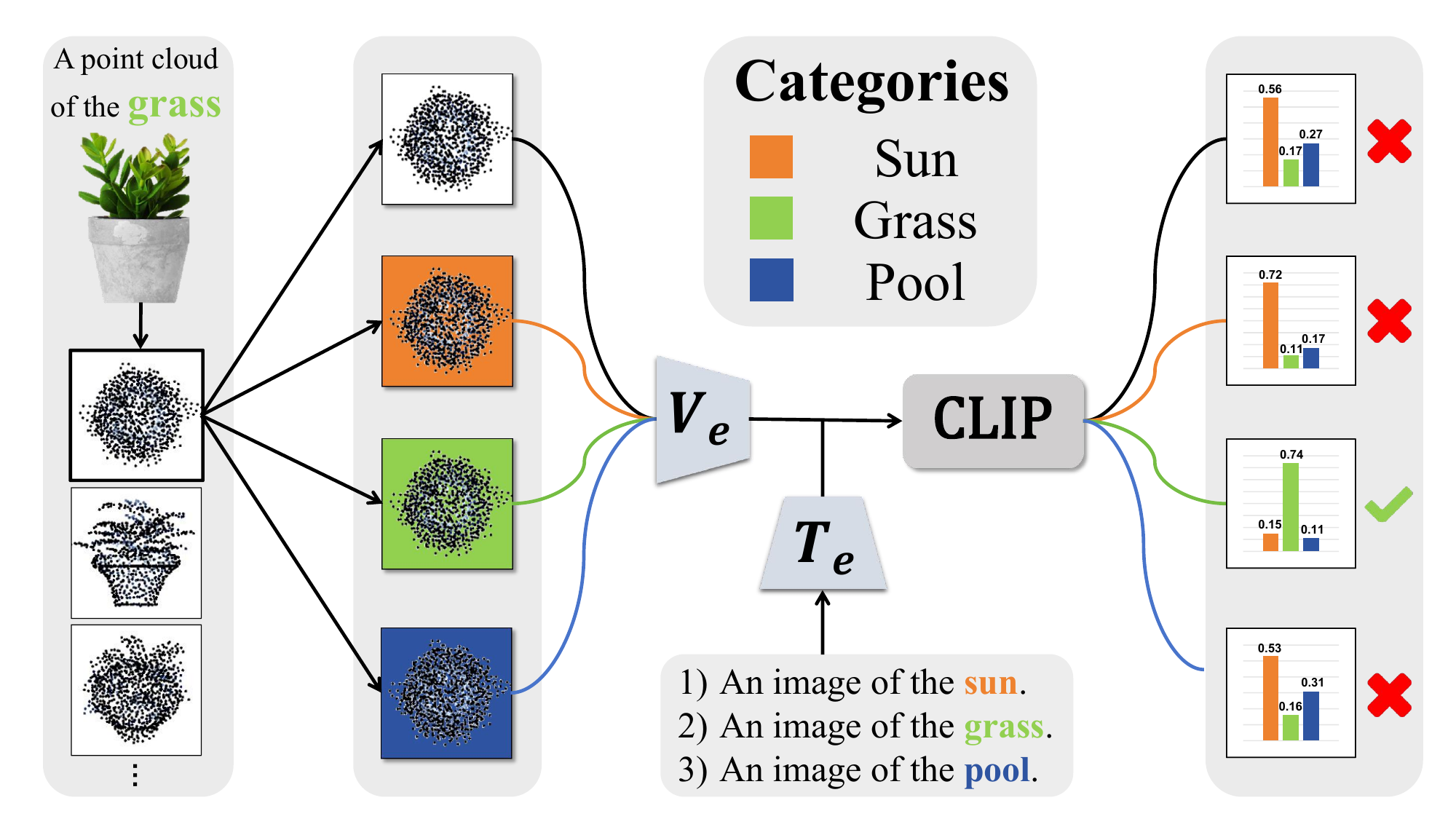}
\caption{
Certain ambiguous projections may inadequately capture the full representation of a given point cloud, and the absence of color information in depth maps introduces inconsistencies with CLIP's input space. To mitigate these issues, we enhance the depth map without altering CLIP's architecture by incorporating a Texture Amplification Module during the rendering process. This approach enables the point cloud to learn an optimal background color, thereby improving the fidelity of auxiliary predictions.}
\label{fig:color}
\end{figure}

While projecting point clouds onto 2D planes has become prevalent for CLIP adaptation, this paradigm faces dual challenges inherited from its projection nature and modality discrepancy. The inherent information loss during 3D-to-2D conversion is exacerbated by viewpoint-sensitive projection consistency, particularly under noisy point cloud conditions. Moreover, the domain gap between CLIP's natural image priors and depth map characteristics fundamentally limits representation alignment, leading to a significant performance drop.

To address these co-occurring challenges without structural overcomplication, we propose the Texture Amplification Module (TAM) that synergizes geometric preservation and modality adaptation. Our key insight stems from the observation that the background color context can guide CLIP's depth map interpretation without architectural modification. As illustrated in Figure~\ref{fig:color}, TAM embeds lightweight learnable color generators within the standard rendering pipeline. For each point cloud $P$, we uniformly sample multiple views $V$ to render depth maps $I^D$. Concurrently, TAM synthesizes adaptive RGB values conditioned on point cloud features $\mathbf{F^P}$ through parameterized transformations:
\begin{equation}
    \mathbf{c} = \frac{\tanh\left( \mathbf{W}_2 \cdot \text{ReLU}(\mathbf{W}_1 \cdot \mathbf{F^P} + \mathbf{b}_1) + \mathbf{b}_2 \right) + 1}{2},
\end{equation}
where $\mathbf{W}_1, \mathbf{W}_2$ and $\mathbf{b}_1, \mathbf{b}_2$ denote learnable parameters, with output $\mathbf{c} \in [0,1]^3$ ensuring valid RGB values. The rendering process strategically applies these colors to mitigate projection artifacts: after identifying white background pixels $M_w$ in $I^D$, we detect kernel-wise background regions via $9\times9$ convolution with padding 4, marking central pixels as $M_b$ when fully surrounded by $M_w$. Filling $M_b$ with learned color $\mathbf{c}$ produces enhanced images $I^E$ while preserving the original depth dimensions.

This dual-stream processing maintains geometric fidelity in $I^D$ while introducing minimal texture cues through $I^E$. The final logits combine geometric features from depth encoder $E_D$ and CLIP's multimodal understanding:
\begin{equation}
    \text{logits} = \mathbf{F^T} \cdot \widehat{\mathbf{F}} + \text{CLIP}(I^E, d),
\end{equation}
where color guidance is optimized via feature alignment loss:
\begin{equation}
    \mathcal{L}_c = \frac{1}{V} \left( \sum_{i=1}^{V} \frac{1 - \cos(\mathbf{F}_i^E, \mathbf{F^T})}{2} \right).
\end{equation}
Classification directly selects the maximal probability from task-specific logits $\mathbb{R}^{B \times C_t}$, eliminating complex decoding steps. The entire framework achieves modality adaptation through learnable color semantics rather than structural intervention, preserving CLIP's original processing flow while compensating for projection deficiencies.

\subsection{Base-Novel Discriminator}

In few-shot class-incremental learning (FSCIL), models are initially trained on a base task with ample data before encountering incremental tasks with only a limited number of samples. This setup enables effective alignment of base classes with their prototypes. However, during incremental learning, training novel classes from scratch is infeasible due to the scarcity of samples. Consequently, novel class training relies on parameters from the base network, introducing disruptions to previously aligned base classes and leading to catastrophic forgetting.

Beyond the ``sample gap'', a domain gap often exists between base and novel classes. Unlike within-domain settings where both sets originate from the same dataset, cross-domain scenarios—where base and novel classes come from different distributions—pose greater challenges. In these cases, the model must generalize effectively in a few-shot setting, complicating feature alignment and unification.

To address these challenges, we introduce the Base-Novel Discriminator (BND), which leverages exemplars to enhance boundary separation and enable parallel processing of base and novel classes. Exemplars, commonly used in class-incremental learning~\cite{rebuffi2017icarl,castro2018end,hou2019learning} and prior 3DFSCIL research~\cite{chowdhury2022few,xu2023filp,tan2024cross,cheraghian2024canonical}, serve as representative samples from learned classes. Given the exemplar set $E_0 = \bigcup_{i=1}^{C_0} X_i$ for the base task, where $C_0$ denotes the number of base classes and $X_i$ contains point cloud exemplars from class $i$. In practice, we limit the number of exemplars per class to 1, i.e., $|X_i|=1$. During each incremental task, we map labels from the base classes to 1 and novel classes to 0, yielding binary labels $y_i$. The BND then performs binary classification on the point cloud feature $F^P$ to produce logits $\hat{y}_i$. The training of BND is conducted independently using binary cross-entropy loss:
\begin{align}
    \mathcal{L}_{\text{BND}} &= \frac{1}{T-1} \sum_{t=1}^{T-1} \frac{1}{|D^t_{\text{train}}|} \sum_{i=1}^{|D^t_{\text{train}}|} \left[ -y_i \log(\hat{y}_i) \right. \nonumber \\
    &\quad \left. - (1 - y_i) \log(1 - \hat{y}_i) \right]
\end{align}
During testing, we perform binary classification based solely on $F^P$. By specifying a threshold $h$, BND maps $\hat{y}_i$ larger than $h$ to base classes and the rest to novel classes. This allows us to process them in parallel. To achieve this, after the training of the base task, we freeze the parameters of the current base network and save it as $Net_B$ for base classes while leaving the original network $Net$ unchanged for the incremental learning of novel classes. Due to the small size of training sets in incremental tasks, BND can be quickly trained independently before each incremental task. It can also dynamically adjust the classification boundary as new classes are introduced. Additionally, its independence from the main network makes it less susceptible to catastrophic forgetting and overfitting.

%% file: sections/4_Experiments.tex
\section{Experiments}

\begin{table*}[!t]
\caption{\small Quantitative comparison under cross-domain settings. The best and second-best results are \textbf{bolded} and \underline{underlined}, respectively.}
\newcolumntype{C}[1]{>{\centering\arraybackslash}p{#1}}
\centering \normalsize 

\scalebox{0.68}{ 
\rowcolors{5}{white!15}{gray!20}
\begin{tabular}{C{2.6cm}|C{0.45cm}C{0.45cm}C{0.45cm}C{0.45cm}C{0.45cm}C{0.45cm}C{0.45cm}C{0.45cm}C{0.45cm}C{0.45cm}C{0.45cm}C{0.45cm}C{0.6cm}|C{0.45cm}C{0.45cm}C{0.45cm}C{0.45cm}C{0.45cm}C{0.6cm}|C{0.45cm}C{0.45cm}C{0.45cm}C{0.45cm}C{0.45cm}C{0.6cm}}
\hline
\multirow{2}{*}{\diagbox{Method}{Task}} & \multicolumn{13}{c|}{ShapeNet $\rightarrow$ CO3D} & \multicolumn{6}{c|}{ ModelNet $\rightarrow$ ScanObjectNN} &\multicolumn{6}{c}{ShapeNet $\rightarrow$ ScanObjectNN}\\
\cline{2-26}
& 39 & 44 & 49 & 54 & 59 & 64 & 69 & 74 & 79 & 84 & 89 & $\Delta_A\downarrow$ & $AA\uparrow$ & 26 & 30 & 34 & 37 & $\Delta_A\downarrow$ & $AA\uparrow$ & 44 & 49 & 54 & 59  & $\Delta_A\downarrow$ & $AA\uparrow$\\
\hline
\textit{FT} & 81.0 & 20.2 & 2.3 & 1.7 & 0.8 & 1.0 & 1.0 & 1.3 & 0.9 & 0.5 & 1.6 & 59.3 & 10.2  & 88.4 & 6.4 & 6.0 & 1.9 & 55.8 & 25.7   & 81.4 & 38.7 & 4.0 & 0.9 & 73.2 & 31.3\\
\textit{Joint} & 81.0 & 79.5 & 78.3 & 75.2 & 75.1 & 74.8 & 72.3 & 71.3 & 70.0 & 68.8 & 67.3 & 1.8 & 74.0 & 88.4 & 79.7 & 74.0 & 71.2 & 6.9 & 78.3 & 81.4 & 82.5 & 79.8 & 78.7 & 2.0 & 80.6\\
\hline
LwF \cite{li2017learning} & 81.0 & 57.4 & 19.3 & 2.3 & 1.0 & 0.9 & 0.8 & 1.3 & 1.1 & 0.8 & 1.9 & 50.4 & 15.3 & 88.4 & 35.8 & 5.8 & 2.5 & 66.7 & 33.1 & 81.4 & 47.9 & 14.0 & 5.9 & 56.6 & 37.3\\
IL2M \cite{belouadah2019il2m} & 81.0 & 45.6 & 36.8 & 35.1 & 31.8 & 33.3 & 34.0 & 31.5 & 30.6 & 32.3 & 30.0 & 10.7 & 38.4 & 88.4 & 58.2 & 52.9 & 52.0 & 15.0 & 62.9 & 81.4 & 53.2 & 43.9 & 45.8 & 18.8 & 56.1\\
ScaIL \cite{belouadah2020scail} & 81.0 & 50.1 & 45.7 & 39.1 & 39.0 & 37.9 & 38.0 & 36.0 & 33.7 & 33.0 & 35.2 & 8.5 & 42.6 & 88.4 & 56.5 & 55.9 & 52.9 & 14.2 & 63.4 & 81.4 & 49.0 & 46.7 & 40.0 & 19.6 & 54.3\\
EEIL \cite{castro2018end} & 81.0 & 75.2 & 69.3 & 63.2 & 60.5 & 57.9 & 53.0 & 51.9 & 51.3 & 47.8 & 47.6 & 5.1 & 59.9 & 88.4 & 70.2 & 61.0 & 56.8 & 13.5 & 69.1 & 81.4 & 74.5 & 69.8 & 63.4 & 8.0 & 72.3\\
FACT \cite{zhou2022forward} & 81.4 & 76.0 & 70.3 & 68.1 & 65.8 & 63.5 & 63.0 & 60.1 & 58.2 & 57.5 & 55.9 & 3.7 & 65.4 & 89.1 & 72.5 & 68.3 & 63.5 & 10.5 & 73.4 & 82.3 & 74.6 & 69.9 & 66.8 & 6.7 & 73.4\\
Sem-aware \cite{cheraghian2021semantic} & 80.6 & 69.5 & 66.5 & 62.9 & 63.2 & 63.0 & 61.2 & 58.3 & 58.1 & 57.2 & 55.2 & 3.7 & 63.2 & 88.5 & {73.9} & 67.7 & 64.2 & 10.0 & 73.6 & 81.3 & 70.6 & 65.2 & 62.9 & 8.1 & 70.0\\
Microshape~\cite{chowdhury2022few}  & {82.6} & {77.9} & {73.9} & {72.7} & {67.7} & {66.2} & {65.4} & {63.4} & {60.6} & {58.1} & {57.1} & {3.6} & 67.8 & 89.3 & 73.2 & {68.4} & {65.1} & {9.8} & 74.0 & {82.5} & {74.8} & {71.2} & {67.1} & 6.6 & 73.9\\
FILP-3D~\cite{xu2023filp}  & {91.4} & \underline{80.7} & \underline{80.6} & \underline{76.2} & \underline{75.7} & {68.2} & {66.8} & {62.9} & {59.1} & {60.2} & {57.1} & {4.9} & 70.8 & \textbf{93.6} & \underline{85.0} & \underline{78.1} & \underline{74.1} & \underline{7.5} & \underline{82.7} & \underline{92.3} & \underline{87.3} & \underline{83.8}& \underline{82.4} & \underline{3.7} & \underline{86.5}\\
C3PR \cite{cheraghian2024canonical} & \underline{83.6} & 80.0 & 77.8 & 75.4 & 72.8 & \underline{72.3} & \underline{70.3} & \underline{67.9} & \underline{64.9} & \underline{64.1} & \underline{63.2} & \textbf{2.8} & \underline{72.0} & 88.3 & 75.7 & 70.6 & 67.8 & 8.3 & 75.6 & 84.5 & 77.8 & 75.5 & 71.9 & 5.2 & 77.4\\
Ours & \textbf{91.4} & \textbf{89.8} & \textbf{86.8} & \textbf{84.3} & \textbf{80.3} & \textbf{78.4} & \textbf{74.9} & \textbf{71.7} & \textbf{71.3} & \textbf{69.0} & \textbf{66.3} & \underline{3.2} & \textbf{78.6} & \underline{93.3}  & \textbf{88.8} & \textbf{81.3} & \textbf{74.9} & \textbf{7.0} & \textbf{84.6} & \textbf{92.4} & \textbf{90.0} & \textbf{87.0} & \textbf{84.0} & \textbf{3.1} & \textbf{88.4}\\
\hline
\end{tabular}

} 
\label{table:cross-domain}
\end{table*}

\begin{table*}[!t]
\centering
\caption{\small Quantitative comparison under within-domain settings. The best and second best results are \textbf{bolded} and \underline{underlined}, respectively.}
\newcolumntype{C}[1]{>{\centering\arraybackslash}p{#1}}
\resizebox{\linewidth}{!}{
\rowcolors{5}{white!15}{gray!20}
\begin{tabular}{C{2.6cm}|C{0.6cm}C{0.6cm}C{0.6cm}C{0.6cm}C{0.6cm}C{0.6cm}C{0.6cm}C{0.6cm}C{0.6cm}|C{0.6cm}C{0.6cm}C{0.6cm}C{0.6cm}C{0.6cm}C{0.6cm}C{0.6cm}}
\hline
\multirow{2}{*}{\diagbox{Method}{Task}} & \multicolumn{9}{c|}{ShapeNet $\rightarrow$ ShapeNet} & \multicolumn{7}{c}{ModelNet $\rightarrow$ ModelNet} \\
\cline{2-17}
& 25 & 30 & 35 & 40 & 45 & 50 & 55 & $\Delta_A\downarrow$ & $AA\uparrow$ & 20 & 25 & 30 & 35 & 40 & $\Delta_A\downarrow$ & $AA\uparrow$ \\
\hline
\textit{FT} & 87.0 & 25.7 & 6.8 & 1.3 & 0.9 & 0.6 & 0.4 & 53.7 & 17.5 & 89.8 & 9.7 & 4.3 & 3.3 & 3.0 & 44.3 & 22.0 \\
\textit{Joint} & 87.0 & 85.2 & 84.3 & 83.0 & 82.5 & 82.2 & 81.3 & 1.1 & 83.6 & 89.8 & 88.2 & 87.0 & 83.5 & 80.5 & 2.7 & 85.8 \\
\hline
LwF \cite{li2017learning} & 87.0 & 60.8 & 33.5 & 15.9 & 3.8 & 3.1 & 1.8 & 44.0 & 29.4 & 89.8 & 36.0 & 9.1 & 3.6 & 3.1 & 52.2 & 28.3 \\
IL2M \cite{belouadah2019il2m} & 87.0 & 58.6 & 45.7 & 40.7 & 50.1 & 49.4 & 49.3 & 15.0 & 54.4 & 89.8 & 65.5 & 58.4 & 52.3 & 53.6 & 12.7 & 63.9 \\
ScaIL \cite{belouadah2020scail} & 87.0 & 56.6 & 51.8 & 44.3 & 50.3 & 46.3 & 45.4 & 13.6 & 54.5 & 89.8 & 66.8 & 64.5 & 58.7 & 56.5 & 10.4 & 67.3 \\
EEIL \cite{castro2018end} & 87.0 & 77.7 & 73.2 & 69.3 & 66.4 & 65.9 & 65.8 & 4.5 & 72.2 & 89.8 & 75.4 & 67.2 & 60.1 & 55.6 & 11.2 & 69.6 \\
FACT \cite{zhou2022forward} & 87.5 & 75.3 & 71.4 & 69.9 & 67.5 & 65.7 & 62.5 & 5.4 & 71.4 & 90.4 & 81.3 & 77.1 & \underline{73.5} & 65.0 & 7.9 & 77.5 \\
Sem-aware \cite{cheraghian2021semantic} & 87.2 & 74.9 & 68.1 & 69.0 & 68.1 & 66.9 & 63.8 & 5.4 & 71.1 & 91.3 & 82.2 & 74.3 & 70.0 & 64.7 & 8.2 & 76.5 \\
Microshape~\cite{chowdhury2022few} & 87.6 & \underline{83.2} & \underline{81.5} & \underline{79.0} & 76.8 & 73.5 & 72.6 & 3.1 & \underline{79.2} & \underline{93.6} & \underline{83.1} & \underline{78.2} & \textbf{75.8} & \underline{67.1} & \underline{7.9} & 79.6 \\
C3PR \cite{cheraghian2024canonical} & \underline{88.0} & 81.6 & 77.8 & 76.7 & \underline{76.9} & \underline{76.2} & \underline{74.7} & \underline{2.7} & 78.8 & 91.6 & 82.3 & 75.8 & 72.2 & \textbf{70.9} & \textbf{6.2} & 78.6 \\
Ours & \textbf{91.6} & \textbf{88.1} & \textbf{87.3} & \textbf{86.3} & \textbf{87.0} & \textbf{86.5} & \textbf{86.4} & \textbf{1.2} & \textbf{87.6} & \textbf{95.0} & \textbf{84.8} & \textbf{81.0} & 72.2 & 65.9 & 8.7 & \textbf{79.8} \\
\hline
\end{tabular}
}
\label{table:within-domain}
\end{table*}

\subsection{Settings}

\noindent\textbf{Datasets.} We evaluate our method on four widely used point cloud benchmarks for experiments, including two synthetic datasets (ShapeNet~\cite{chang2015shapenet} and ModelNet~\cite{wu20153d}) and two real scanned datasets (CO3D~\cite{reizenstein2021common} and ScanObjectNN~\cite{uy2019revisiting}). For a thorough evaluation, we consider two few-shot incremental learning settings: 
\begin{itemize}[leftmargin=*]
    \item \textbf{Cross-domain 3DFSCIL.} To mimic realistic scenarios where only synthetic data are available for training models that must eventually generalize to in-the-wild object recognition, we perform three cross-domain experiments: ShapeNet$\to$CO3D~(S2C), ModelNet$\to$ScanObjectNN (M2O), and ShapeNet$\to$ScanObjectNN (S2O). In each case, we select base training categories from the source dataset that do not overlap with those in the target dataset. The ratios of base to novel classes are 39/50, 26/11, and 44/11 for S2C, M2O, and S2O, respectively. For S2C, the novel classes are divided into 10 incremental tasks (5 classes each). In the M2O setting, the first two incremental tasks comprise 4 novel classes apiece, followed by a final task with 3 classes. For S2O, 15 novel classes from ScanObjectNN are allocated evenly into 3 incremental tasks, with 5 classes per task.
    \item \textbf{Within-domain 3DFSCIL.} Following the protocol of previous studies~\cite{cheraghian2024canonical}, we also conduct within-domain experiments on two synthetic-to-synthetic scenarios: ShapeNet$\to$ShapeNet (S2S) and ModelNet$\to$ModelNet (M2M). For ShapeNet, we designate the 25 categories with the largest sample sizes as base classes and uniformly partition the remaining categories into 6 incremental tasks. Similarly, for ModelNet, 20 base classes are chosen, with the remaining categories divided into 4 incremental tasks, each comprising 5 novel classes.
\end{itemize}

\noindent\textbf{Compared Methods.} We evaluate our approach against contemporary FSCIL techniques designed for both 2D and 3D contexts. The 2D FSCIL methods include LwF~\cite{li2017learning}, IL2M~\cite{belouadah2019il2m}, ScaIL~\cite{belouadah2020scail}, EEIL~\cite{castro2018end}, FACT~\cite{zhou2022forward}, and Sem-aware~\cite{cheraghian2021semantic}, while the 3D FSCIL methods encompass Microshape~\cite{chowdhury2022few}, FILP-3D~\cite{xu2023filp}, and C3PR~\cite{cheraghian2024canonical}. To facilitate a direct comparison with the 3D methods, we replace the 2D backbones of 2D methods with the same 3D encoder as implemented in C3PR~\cite{cheraghian2024canonical}, preserving their original training paradigms. For a fair comparison, we directly adopt the results reported in C3PR~\cite{cheraghian2024canonical}, except for FILP-3D. Since the incremental settings for FILP-3D differ slightly in terms of the number of base and novel categories, we reproduce their results under the same conditions using their official code. 

\noindent\textbf{Metrics.} Following C3PR~\cite{cheraghian2024canonical}, we utilize micro-accuracy across all available categories to assess the performance of our method at each incremental task. Additionally, we report the average accuracy across all incremental tasks, referred to as \textit{AA}. To assess the model's resilience to catastrophic forgetting, we compute the averaged relative performance degradation, defined as $ \Delta_A = \frac{1}{T-1} \sum_{t=0}^{T-2} \frac{|Acc_{t} - Acc_{t+1}|}{Acc_t}$, where $Acc_{t}$ represents the accuracy at task $t$, $t=0$ for the base task, and $T$ denotes the number of tasks contained in current setting.

\renewcommand{\arraystretch}{1.2}
\begin{table*}[!t]
\centering
\caption{\small Ablation studies on each module of our proposed framework on ShapeNet$\to$CO3D.}
\newcolumntype{C}[1]{>{\centering\arraybackslash}p{#1}}
\resizebox{\linewidth}{!}{
\rowcolors{2}{white!15}{gray!20}
\begin{tabular}{C{1.2cm}|C{0.7cm}C{0.7cm}C{0.7cm}|C{0.6cm}C{0.6cm}C{0.6cm}C{0.6cm}C{0.6cm}C{0.6cm}C{0.6cm}C{0.6cm}C{0.6cm}C{0.6cm}C{0.6cm}|C{0.7cm}C{0.7cm}C{0.7cm}}
\hline
Variants & SAGR & TAM & BND & 39 & 44 & 49 & 54 & 59 & 64 & 69 & 74 & 79 & 84 & 89 & $AA\uparrow$ & PD(\%) & $\Delta_A\downarrow$\\
\hline
Baseline &   &   &                   & 91.2 & 88.1 & 85.5 & 81.3 & 76.2 & 72.8 & 65.6 & 62.8 & 62.1 & 58.2 & 51.9 & 72.3 & 6.3 & 5.4 \\
V1 & \ding{51}  &   &          & 91.3 & 88.8 & 84.8 & 82.0 & 78.1 & 74.2 & 69.5 & 65.1 & 64.6 & 61.2 & 57.7 & 74.3 & 4.3 & 4.5 \\
V2 &  & \ding{51} &            & 91.3 & 88.8 & 85.1 & 81.7 & 76.1 & 72.3 & 69.1 & 63.9 & 62.7 & 57.0 & 53.8 & 72.9 & 5.7 & 5.1 \\
V3 & &   &   \ding{51}         & 91.4 & 88.8 & 85.8 & 81.5 & 77.6 & 74.6 & 68.6 & 65.9 & 64.7 & 59.9 & 58.2 & 74.3 & 4.3 & 4.4 \\
V4 & & \ding{51} & \ding{51}  & 91.4 & 88.7 & 86.3 & 82.1 & 78.1 & 74.9 & 69.9 & 66.8 & 65.9 & 62.1 & 59.8 & 75.1 & 3.5 & 4.1 \\
V5 & \ding{51} &   & \ding{51} & 91.3 & 87.2 & 86.4 & 81.5 & 78.8 & 75.0 & 72.5 & 67.5 & 69.4 & 66.5 & 64.0 & 76.4 & 2.2 & 4.0 \\
V6 & \ding{51}  & \ding{51} &  & 91.2 & 88.9 & 84.7 & 81.2 & 77.5 & 74.7 & 70.8 & 67.9 & 67.2 & 63.1 & 60.1 & 75.3 & 3.3 & 4.1 \\
Full & \ding{51} & \ding{51} & \ding{51} & \textbf{91.4} & \textbf{89.8} & \textbf{86.8} & \textbf{84.3} & \textbf{80.3} & \textbf{78.4} & \textbf{74.9} & \textbf{71.7} & \textbf{71.3} & \textbf{69.0} & \textbf{66.3} & \textbf{78.6} & - & \textbf{3.2} \\
\hline
\end{tabular}
}
\label{table:ablations}
\end{table*}

\noindent\textbf{Implementation Details.}
We utilize Recon~\cite{qi2023contrast} as the point cloud feature extractor. Consistent with prior work, the depth encoder and 2D foundation model are configured as in FILP-3D~\cite{xu2023filp}. Specifically, we employ the depth model from~\cite{huang2023clip2point} and the pretrained CLIP ViT-B/32. Training is conducted using the Adam optimizer~\cite{kingma2014adam} with a weight decay of 1e-4, with 10 epochs for base task training and 20 epochs for incremental tasks. For each incremental task, the learning rate is initialized at 1e-3 and decays to 1e-4 according to a cosine annealing schedule. Additionally, the Base-Novel Discriminator is trained separately for 10 epochs at a learning rate of 1e-3 before each incremental update. The mask ratio $M_R$ and the number of self-attention layers $N_{sa}$ in SAGR are set to 0.9 and 2, respectively. While the selected layers $L_r$ for performing geometric rectification are set to \{0, 4, 8\}, the threshold $h$ for BND is set as 0.1, respectively. All experiments are performed on a single NVIDIA RTX3090 GPU with a batch size of 16. 

\subsection{Quantitative Comparison}

The quantitative results for both cross-domain and within-domain 3DFSCIL are presented in Table~\ref{table:cross-domain} and Table~\ref{table:within-domain}, respectively. In addition to the comparison of various methods, we provide baseline results for fine-tuning as a rough benchmark to establish the lower and upper bounds. Specifically, $FT$ (Fine-Tuning) refers to the approach where the model is initialized with the weights from the previous task and further fine-tuned using only the limited samples from the new incremental task. On the other hand, $Joint$ represents the model trained on all available samples from the new classes in each incremental task.

As shown in Table~\ref{table:cross-domain}, our method significantly outperforms existing techniques across all three cross-domain 3DFSCIL tasks. Traditional 2D-based methods experience catastrophic performance degradation when naively adapted to 3D via simple encoder replacement. This underscores the critical flaw in relying exclusively on 2D, texture-biased features for capturing the intricacies of 3D geometric hierarchies. While both C3PR~\cite{cheraghian2024canonical} and FILP-3D~\cite{xu2023filp} mitigate catastrophic forgetting using CLIP-enhanced strategies, their single-stage 3D-to-2D projection suffers from indiscriminate fusion of geometric and textural cues. Notably, in the ShapeNet$\to$CO3D cross-domain scenario, our method achieves a final-task accuracy of 66.3\%, outperforming FILP-3D’s 57.1\% with a relatively lower averaged forgetting rate of 3.2\%, demonstrating the critical role of leveraging intermediate spatial priors for enhanced cross-domain geometric generalization.

Furthermore, as shown in Table~\ref{table:within-domain}, we also present results for the model's performance under two within-domain settings: S2S and M2M. In contrast to cross-domain settings, within-domain tasks are more focused on detecting novel classes within a known distribution rather than entirely new, unseen anomalies. The results across both within-domain settings further support the efficacy of our proposed framework, as our method consistently outperforms existing approaches, demonstrating its robustness in scenarios where new classes emerge from within the same distribution.


\subsection{Ablation Studies}

\noindent\textbf{Ablation of Sub-modules.}
We designed six variants to evaluate the contribution of each module in the model by isolating some of the components. The overall results are shown in Table \ref{table:ablations}, where $PD$ stands for Performance Drop here. Compared to the baseline, our proposed framework achieves 6.3\% improvement on the overall accuracy. The baseline here refers to the CMGR variant without SARG, TAM, and BND under the 3DFSCIL setting.
V1, V2, and V3 respectively show the impact of introducing SAGR, TAM, and BND individually on performance. Since SAGR alleviates the semantic ambiguity between 3D information and 2D prior contained in encoders through cross-modal geometric rectification, it enhances the model's ability to represent point clouds, resulting in a significant performance improvement. Additionally, TAM helps mitigate the domain gap between depth maps and RGB images by aligning depth maps with CLIP’s input space, thereby assisting the prediction. Furthermore, to maintain geometric independence between base and novel classes, BND performs a preliminary binary classification, which greatly alleviates the problem of catastrophic forgetting. V4, V5, and V6 present the impact of single-module ablation or different module combinations on performance. Results indicate that the absence of any module negatively affects performance, while the combination of multiple modules outperforms any single module used individually.

\begin{figure}
\centering
\includegraphics[width=1\columnwidth]{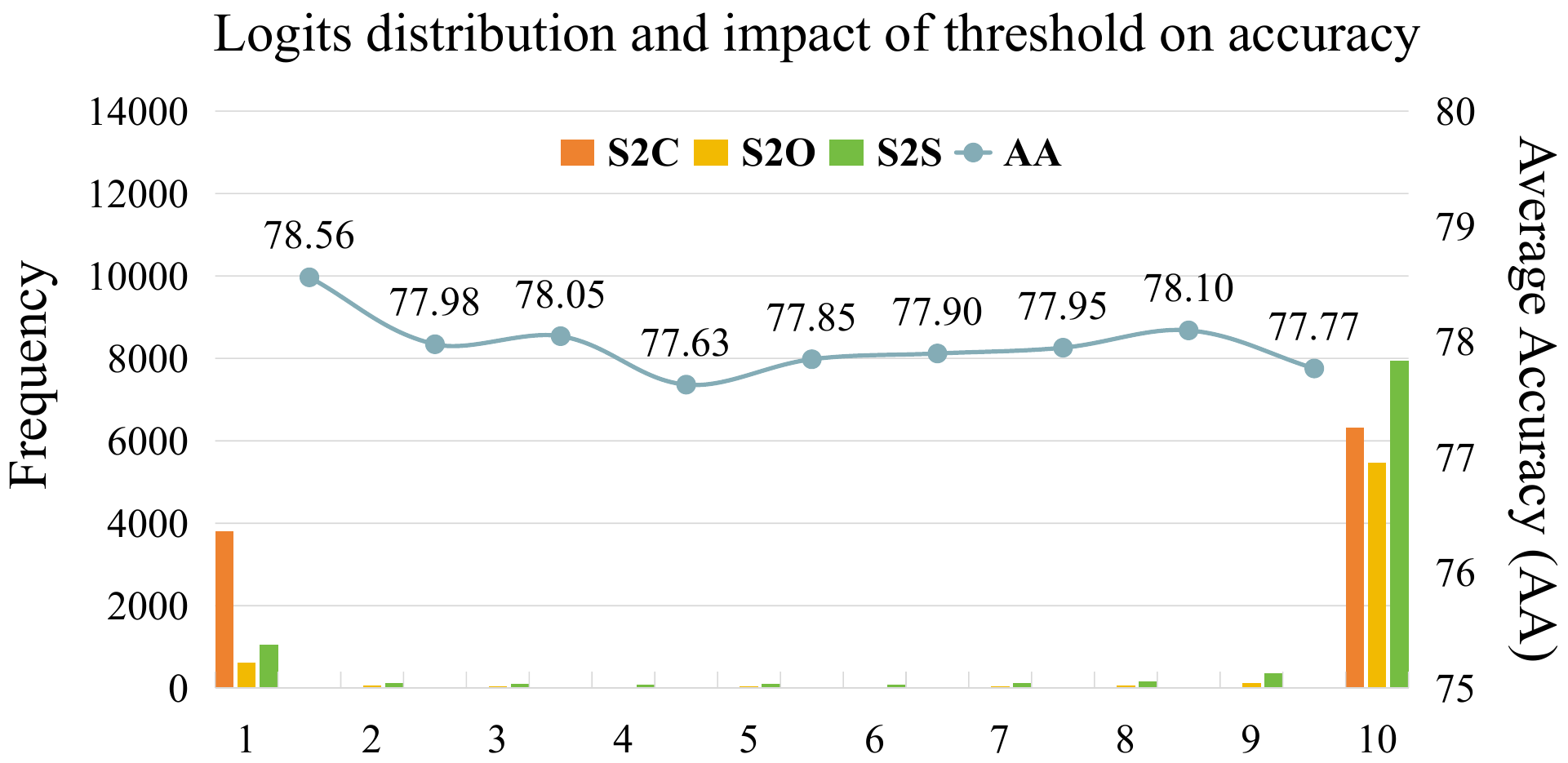}
\caption{
Base-Novel Discriminator can effectively distinguish between base and novel classes while remaining robust to hyperparameter variations. On the horizontal axis, "1" represents a logit range of 0 to 0.1, and so forth. The plotted curve illustrates the accuracy when the binary classification threshold is set to the upper bound of each interval.}
\vspace{-4mm}
\label{fig:thres_logits}
\end{figure}

\noindent\textbf{Sensitivity Analysis for Hyperparameters.} 
We investigate how the predefined threshold $h$ in the Base-Novel Discriminator affects the final performance on the S2C setting. As presented in Figure \ref{fig:thres_logits}, the output logits are well separated to the extreme value of high confidence, which offers a more relaxed region for the selection of $h$.
Furthermore, we investigate the impact of the prior knowledge contained in different CLIP layers on model performance. More details of the selection of hyperparameters can be found in the supplementary materials.

\subsection{Visualizations}
\begin{figure}
\centering
\includegraphics[width=1\columnwidth]{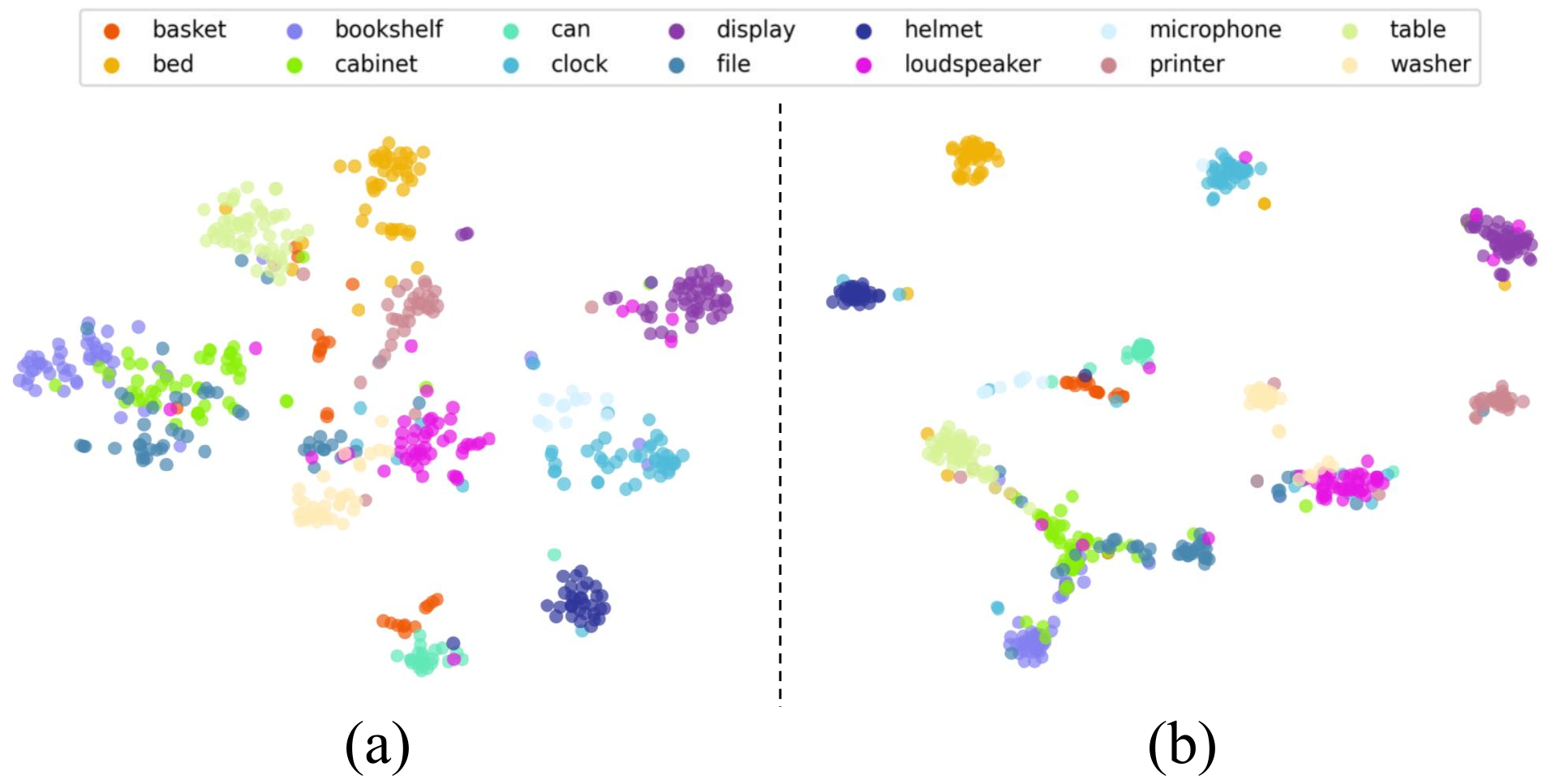}
\caption{T-SNE visualizations before and after employing CMGR. (a) Relying solely on unrectified 3D features leads to sparse intra-class distributions and unclear inter-class boundaries, while (b) incorporating geometric rectification significantly enhances the compactness and discriminability of the feature space.}
\label{fig:tsne}
\end{figure}

We present additional visualization results to provide a more intuitive understanding of the effectiveness of our proposed method. Figure~\ref{fig:tsne} illustrates the feature space before and after the application of cross-modal geometric rectification. As depicted in Figure~\ref{fig:tsne}(a), when the point cloud features are solely processed by the 3D point encoder, the feature space is disorganized, largely due to the fine-grained semantic variations inherent within the same category (e.g., the airplane class encompassing subtypes such as fighter jets and passenger planes). This causes the extracted features to become biased, leading to sparse intra-class distributions. Furthermore, geometric ambiguities across different categories (e.g., tables and chairs sharing similar characteristics, such as four legs and a flat surface) exacerbate the issue, reducing inter-class discriminability.
The feature space after the application of cross-modal geometric rectification is shown in Figure~\ref{fig:tsne}(b). By introducing the Structure-Aware Geometric Rectification (SAGR) module, which hierarchically integrates semantic information from 2D data, our model achieves a more compact and distinctive representation of point clouds. This reduction in feature ambiguity not only facilitates more effective incremental learning of novel categories but also preserves the discriminability of historical tasks, ensuring better overall performance.

\begin{figure}
\centering
\includegraphics[width=\columnwidth]{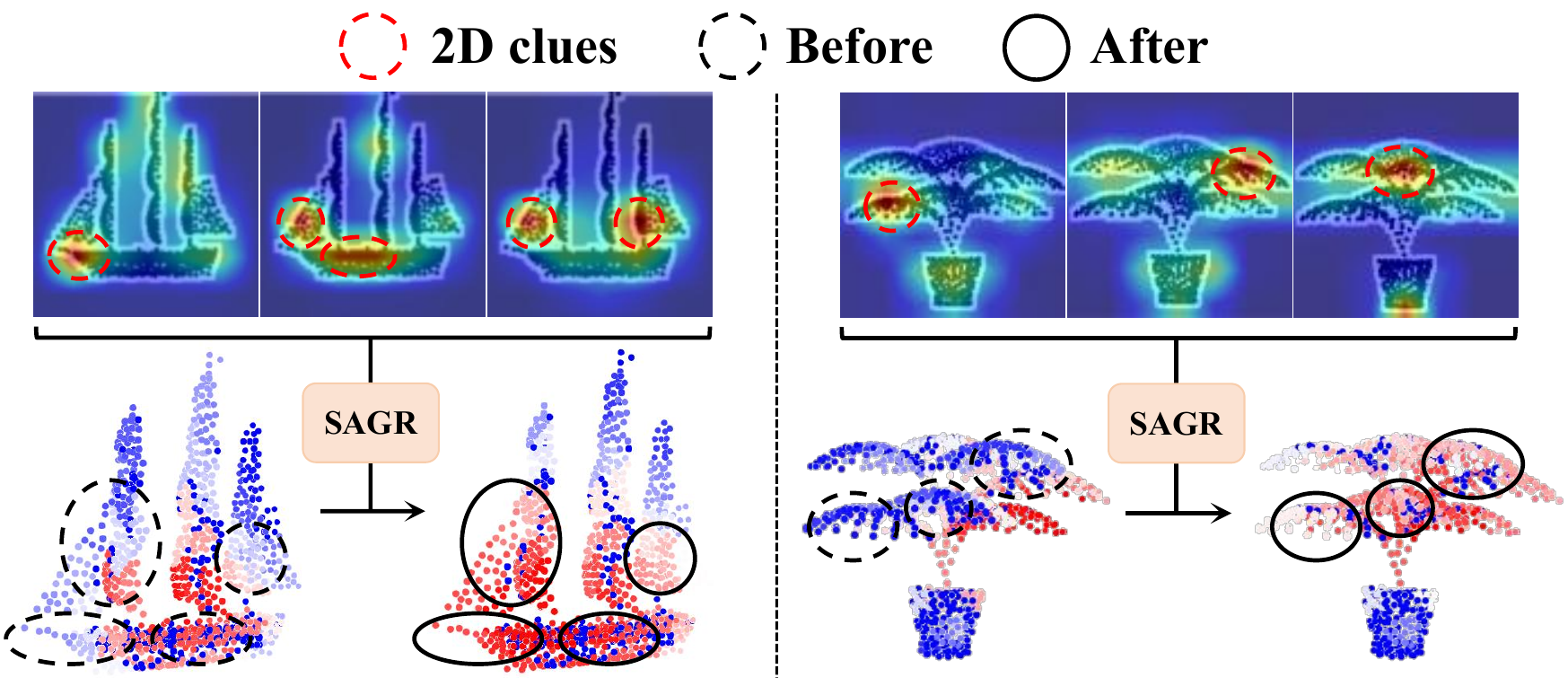}
\caption{Point cloud attentions before and after geometric rectification, the red regions have higher attention weights compared to the blue regions.
}
\vspace{-3mm}
\label{fig:reb_sagr_Yaxis}
\end{figure}

Additionally, in Figure~\ref{fig:reb_sagr_Yaxis}, we visualize the point features processed by the SAGR module, where red regions indicate higher weights (i.e., receiving more attention), while blue regions indicate lower weights (i.e., receiving less attention). Compared to directly using the point encoder to extract point cloud features, point clouds rectified via 2D information exhibit a broader attention distribution. In particular, the ability to capture local features is enhanced, leading to improved distinguishability between samples.




%% file: sections/5_Conclusion.tex
\section{Conclusion}
This work tackles geometric misalignment and texture bias in 3DFSCIL through cross-modal geometric rectification, integrating CLIP’s intermediate spatial semantics to enhance 3D representations. By leveraging CLIP’s spatial priors, our approach mitigates semantic blurring from texture-biased projections and indiscriminate fusion. The Structure-Aware Geometric Rectification module aligns 3D part hierarchies with CLIP’s spatial relationships, while the Texture Amplification Module refines textures to suppress noise and reinforce consistency. A Base-Novel Discriminator stabilizes incremental prototypes, preventing catastrophic forgetting. Experiments confirm state-of-the-art performance, redefining CLIP as a geometry-refining lens for 3D learning under extreme data scarcity.
While our framework achieves strong cross-modal alignment, it assumes that CLIP’s spatial priors sufficiently represent 3D structures, which may not generalize to extreme cases like highly deformable or occluded objects. Future work will explore adaptive geometric priors and uncertainty modeling to improve robustness in complex 3D scenarios.

\noindent\textbf{Acknowledgement.} 
The work is supported by Guangdong Provincial Natural Science Foundation for Outstanding Youth Team Project (No.: 2024B1515040010), NSFC Key Project (No.: U23A20391), China National Key R\&D Program (No.: 2023YFE0202700, 2024YFB4709200), Key-Area Research and Development Program of Guangzhou City (No.: 2023B01J0022), Postdoctoral Fellowship Program (Grade C) of China Postdoctoral Science Foundation (No.: GZC20240498), the Guangdong Natural Science Funds for Distinguished Young Scholars (No.: 2023B1515020097), the National Research Foundation, Singapore under its AI Singapore Programme (No.: AISG3-GV-2023-011), the Singapore Ministry of Education AcRF Tier 1 Grant (No.: MSS25C004), and the Lee Kong Chian Fellowships.

%% file: egpaper_for_review.bbl
\begin{thebibliography}{10}\itemsep=-1pt

\bibitem{afham2022crosspoint}
Mohamed Afham, Isuru Dissanayake, Dinithi Dissanayake, Amaya Dharmasiri, Kanchana Thilakarathna, and Ranga Rodrigo.
\newblock Crosspoint: Self-supervised cross-modal contrastive learning for 3d point cloud understanding.
\newblock In {\em Proceedings of the IEEE/CVF Conference on Computer Vision and Pattern Recognition}, pages 9902--9912, 2022.

\bibitem{agarwal2022semantics}
Aishwarya Agarwal, Biplab Banerjee, Fabio Cuzzolin, and Subhasis Chaudhuri.
\newblock Semantics-driven generative replay for few-shot class incremental learning.
\newblock In {\em Proceedings of the ACM International Conference on Multimedia}, pages 5246--5254, 2022.

\bibitem{belouadah2019il2m}
Eden Belouadah and Adrian Popescu.
\newblock Il2m: Class incremental learning with dual memory.
\newblock In {\em Proceedings of the IEEE/CVF International Conference on Computer Vision}, pages 583--592, 2019.

\bibitem{belouadah2020scail}
Eden Belouadah and Adrian Popescu.
\newblock Scail: Classifier weights scaling for class incremental learning.
\newblock In {\em Proceedings of the IEEE/CVF Winter Conference on Applications of Computer Vision}, pages 1266--1275, 2020.

\bibitem{mann2020language}
Tom Brown, Benjamin Mann, Nick Ryder, Melanie Subbiah, Jared~D Kaplan, Prafulla Dhariwal, Arvind Neelakantan, Pranav Shyam, Girish Sastry, Amanda Askell, et~al.
\newblock Language models are few-shot learners.
\newblock {\em Advances in Neural Information Processing Systems}, 33:1877--1901, 2020.

\bibitem{castro2018end}
Francisco~M Castro, Manuel~J Mar{\'\i}n-Jim{\'e}nez, Nicol{\'a}s Guil, Cordelia Schmid, and Karteek Alahari.
\newblock End-to-end incremental learning.
\newblock In {\em European Conference on Computer Vision}, pages 233--248, 2018.

\bibitem{chang2015shapenet}
Angel~X Chang, Thomas Funkhouser, Leonidas Guibas, Pat Hanrahan, Qixing Huang, Zimo Li, Silvio Savarese, Manolis Savva, Shuran Song, Hao Su, et~al.
\newblock Shapenet: An information-rich 3d model repository.
\newblock {\em arXiv preprint arXiv:1512.03012}, 2015.

\bibitem{cheraghian2024canonical}
Ali Cheraghian, Zeeshan Hayder, Sameera Ramasinghe, Shafin Rahman, Javad Jafaryahya, Lars Petersson, and Mehrtash Harandi.
\newblock Canonical shape projection is all you need for 3d few-shot class incremental learning.
\newblock In {\em European Conference on Computer Vision}. Springer, 2024.

\bibitem{cheraghian2021semantic}
Ali Cheraghian, Shafin Rahman, Pengfei Fang, Soumava~Kumar Roy, Lars Petersson, and Mehrtash Harandi.
\newblock Semantic-aware knowledge distillation for few-shot class-incremental learning.
\newblock In {\em Proceedings of the IEEE/CVF Conference on Computer Vision and Pattern Recognition}, pages 2534--2543, 2021.

\bibitem{chi2022metafscil}
Zhixiang Chi, Li Gu, Huan Liu, Yang Wang, Yuanhao Yu, and Jin Tang.
\newblock Metafscil: A meta-learning approach for few-shot class incremental learning.
\newblock In {\em Proceedings of the IEEE/CVF Conference on Computer Vision and Pattern Recognition}, pages 14166--14175, 2022.

\bibitem{chowdhury2022few}
Townim Chowdhury, Ali Cheraghian, Sameera Ramasinghe, Sahar Ahmadi, Morteza Saberi, and Shafin Rahman.
\newblock Few-shot class-incremental learning for 3d point cloud objects.
\newblock In {\em European Conference on Computer Vision}, pages 204--220. Springer, 2022.

\bibitem{deitke2023objaverse}
Matt Deitke, Dustin Schwenk, Jordi Salvador, Luca Weihs, Oscar Michel, Eli VanderBilt, Ludwig Schmidt, Kiana Ehsani, Aniruddha Kembhavi, and Ali Farhadi.
\newblock Objaverse: A universe of annotated 3d objects.
\newblock In {\em Proceedings of the IEEE/CVF Conference on Computer Vision and Pattern Recognition}, pages 13142--13153, 2023.

\bibitem{devlin2019bert}
Jacob Devlin, Ming-Wei Chang, Kenton Lee, and Kristina Toutanova.
\newblock Bert: Pre-training of deep bidirectional transformers for language understanding.
\newblock In {\em Proceedings of NAACL-HLT}, pages 4171--4186, 2019.

\bibitem{dong2021i3dol}
Jiahua Dong, Yang Cong, Gan Sun, Bingtao Ma, and Lichen Wang.
\newblock I3dol: Incremental 3d object learning without catastrophic forgetting.
\newblock In {\em Proceedings of the Annual AAAI Conference on Artificial Intelligence}, volume~35, pages 6066--6074, 2021.

\bibitem{hou2019learning}
Saihui Hou, Xinyu Pan, Chen~Change Loy, Zilei Wang, and Dahua Lin.
\newblock Learning a unified classifier incrementally via rebalancing.
\newblock In {\em Proceedings of the IEEE/CVF Conference on Computer Vision and Pattern Recognition}, pages 831--839, 2019.

\bibitem{huang2023clip2point}
Tianyu Huang, Bowen Dong, Yunhan Yang, Xiaoshui Huang, Rynson~WH Lau, Wanli Ouyang, and Wangmeng Zuo.
\newblock Clip2point: Transfer clip to point cloud classification with image-depth pre-training.
\newblock In {\em Proceedings of the IEEE/CVF International Conference on Computer Vision}, pages 22157--22167, 2023.

\bibitem{kingma2014adam}
Diederik~P Kingma and Jimmy Ba.
\newblock Adam: A method for stochastic optimization.
\newblock {\em arXiv preprint arXiv:1412.6980}, 2014.

\bibitem{kukleva2021generalized}
Anna Kukleva, Hilde Kuehne, and Bernt Schiele.
\newblock Generalized and incremental few-shot learning by explicit learning and calibration without forgetting.
\newblock In {\em Proceedings of the IEEE/CVF International Conference on Computer Vision}, pages 9020--9029, 2021.

\bibitem{li2017learning}
Zhizhong Li and Derek Hoiem.
\newblock Learning without forgetting.
\newblock {\em IEEE Transactions on Pattern Analysis and Machine Intelligence}, 40(12):2935--2947, 2017.

\bibitem{liu2025genpoly}
Bangzhen Liu, Yu yuyang, Xuemiao Xu, Cheng Xu, Chenxi Zheng, and Shengfeng He.
\newblock Genpoly: Learning generalized and tessellated shape priors via 3d polymorphic evolving.
\newblock {\em IEEE Transactions on Pattern Analysis and Machine Intelligence}, 2025.

\bibitem{liu2025rotation}
Bangzhen Liu, Chenxi Zheng, Xuemiao Xu, Cheng Xu, Huaidong Zhang, and Shengfeng He.
\newblock Rotation-adaptive point cloud domain generalization via intricate orientation learning.
\newblock {\em IEEE Transactions on Pattern Analysis and Machine Intelligence}, 2025.

\bibitem{liu2022few}
Huan Liu, Li Gu, Zhixiang Chi, Yang Wang, Yuanhao Yu, Jun Chen, and Jin Tang.
\newblock Few-shot class-incremental learning via entropy-regularized data-free replay.
\newblock In {\em European Conference on Computer Vision}, pages 146--162. Springer, 2022.

\bibitem{lu20243d}
Yuqin Lu, Bailin Deng, Zhixuan Zhong, Tianle Zhang, Yuhui Quan, Hongmin Cai, and Shengfeng He.
\newblock 3d snapshot: Invertible embedding of 3d neural representations in a single image.
\newblock {\em IEEE Transactions on Pattern Analysis and Machine Intelligence}, 46(12):11524--11531, 2024.

\bibitem{peng2022few}
Can Peng, Kun Zhao, Tianren Wang, Meng Li, and Brian~C Lovell.
\newblock Few-shot class-incremental learning from an open-set perspective.
\newblock In {\em European Conference on Computer Vision}, pages 382--397. Springer, 2022.

\bibitem{qi2017pointnet}
Charles~R Qi, Hao Su, Kaichun Mo, and Leonidas~J Guibas.
\newblock Pointnet: Deep learning on point sets for 3d classification and segmentation.
\newblock In {\em Proceedings of the IEEE/CVF Conference on Computer Vision and Pattern Recognition}, pages 652--660, 2017.

\bibitem{qi2017pointnet++}
Charles~Ruizhongtai Qi, Li Yi, Hao Su, and Leonidas~J Guibas.
\newblock Pointnet++: Deep hierarchical feature learning on point sets in a metric space.
\newblock {\em Advances in Neural Information Processing Systems}, 30, 2017.

\bibitem{qi2023contrast}
Zekun Qi, Runpei Dong, Guofan Fan, Zheng Ge, Xiangyu Zhang, Kaisheng Ma, and Li Yi.
\newblock Contrast with reconstruct: Contrastive 3d representation learning guided by generative pretraining.
\newblock In {\em International Conference on Machine Learning}, pages 28223--28243. PMLR, 2023.

\bibitem{radford2021learning}
Alec Radford, Jong~Wook Kim, Chris Hallacy, Aditya Ramesh, Gabriel Goh, Sandhini Agarwal, Girish Sastry, Amanda Askell, Pamela Mishkin, Jack Clark, et~al.
\newblock Learning transferable visual models from natural language supervision.
\newblock In {\em International Conference on Machine Learning}, pages 8748--8763. PMLR, 2021.

\bibitem{rebuffi2017icarl}
Sylvestre-Alvise Rebuffi, Alexander Kolesnikov, Georg Sperl, and Christoph~H Lampert.
\newblock icarl: Incremental classifier and representation learning.
\newblock In {\em Proceedings of the IEEE/CVF Conference on Computer Vision and Pattern Recognition}, pages 2001--2010, 2017.

\bibitem{reizenstein2021common}
Jeremy Reizenstein, Roman Shapovalov, Philipp Henzler, Luca Sbordone, Patrick Labatut, and David Novotny.
\newblock Common objects in 3d: Large-scale learning and evaluation of real-life 3d category reconstruction.
\newblock In {\em Proceedings of the IEEE/CVF International Conference on Computer Vision}, pages 10901--10911, 2021.

\bibitem{tan2024cross}
Yuwen Tan and Xiang Xiang.
\newblock Cross-domain few-shot incremental learning for point-cloud recognition.
\newblock In {\em Proceedings of the IEEE/CVF Winter Conference on Applications of Computer Vision}, pages 2307--2316, 2024.

\bibitem{tao2020few}
Xiaoyu Tao, Xiaopeng Hong, Xinyuan Chang, Songlin Dong, Xing Wei, and Yihong Gong.
\newblock Few-shot class-incremental learning.
\newblock In {\em Proceedings of the IEEE/CVF Conference on Computer Vision and Pattern Recognition}, pages 12183--12192, 2020.

\bibitem{uy2019revisiting}
Mikaela~Angelina Uy, Quang-Hieu Pham, Binh-Son Hua, Thanh Nguyen, and Sai-Kit Yeung.
\newblock Revisiting point cloud classification: A new benchmark dataset and classification model on real-world data.
\newblock In {\em Proceedings of the IEEE/CVF International Conference on Computer Vision}, pages 1588--1597, 2019.

\bibitem{wang2019dynamic}
Yue Wang, Yongbin Sun, Ziwei Liu, Sanjay~E Sarma, Michael~M Bronstein, and Justin~M Solomon.
\newblock Dynamic graph cnn for learning on point clouds.
\newblock {\em ACM Transactions on Graphics}, 38(5):1--12, 2019.

\bibitem{wu20153d}
Zhirong Wu, Shuran Song, Aditya Khosla, Fisher Yu, Linguang Zhang, Xiaoou Tang, and Jianxiong Xiao.
\newblock 3d shapenets: A deep representation for volumetric shapes.
\newblock In {\em Proceedings of the IEEE/CVF Conference on Computer Vision and Pattern Recognition}, pages 1912--1920, 2015.

\bibitem{xu2025fr2seg}
Cheng Xu, Weiwen Zhang, Hongrui Zhang, Xuemiao Xu, Huaidong Zhang, Jing Zou, and Jing Qin.
\newblock Fr$^2$seg: Continual segmentation across multiple sites via fourier style replay and adaptive consistency regularization.
\newblock In {\em Proceedings of the Annual AAAI Conference on Artificial Intelligence}, volume~39, pages 8815--8823, 2025.

\bibitem{xu2023filp}
Wan Xu, Tianyu Huang, Tianyu Qu, Guanglei Yang, Yiwen Guo, and Wangmeng Zuo.
\newblock Filp-3d: Enhancing 3d few-shot class-incremental learning with pre-trained vision-language models.
\newblock {\em arXiv preprint arXiv:2312.17051}, 2023.

\bibitem{xu2024learning}
Yingjie Xu, Bangzhen Liu, Hao Tang, Bailin Deng, and Shengfeng He.
\newblock Learning with unreliability: fast few-shot voxel radiance fields with relative geometric consistency.
\newblock In {\em Proceedings of the IEEE/CVF Conference on Computer Vision and Pattern Recognition}, pages 20342--20351, 2024.

\bibitem{xue2023ulip}
Le Xue, Mingfei Gao, Chen Xing, Roberto Mart{\'\i}n-Mart{\'\i}n, Jiajun Wu, Caiming Xiong, Ran Xu, Juan~Carlos Niebles, and Silvio Savarese.
\newblock Ulip: Learning a unified representation of language, images, and point clouds for 3d understanding.
\newblock In {\em Proceedings of the IEEE/CVF Conference on Computer Vision and Pattern Recognition}, pages 1179--1189, 2023.

\bibitem{zhang2021few}
Chi Zhang, Nan Song, Guosheng Lin, Yun Zheng, Pan Pan, and Yinghui Xu.
\newblock Few-shot incremental learning with continually evolved classifiers.
\newblock In {\em Proceedings of the IEEE/CVF Conference on Computer Vision and Pattern Recognition}, pages 12455--12464, 2021.

\bibitem{zhang2022pointclip}
Renrui Zhang, Ziyu Guo, Wei Zhang, Kunchang Li, Xupeng Miao, Bin Cui, Yu Qiao, Peng Gao, and Hongsheng Li.
\newblock Pointclip: Point cloud understanding by clip.
\newblock In {\em Proceedings of the IEEE/CVF Conference on Computer Vision and Pattern Recognition}, pages 8552--8562, 2022.

\bibitem{zhao2023few}
Linglan Zhao, Jing Lu, Yunlu Xu, Zhanzhan Cheng, Dashan Guo, Yi Niu, and Xiangzhong Fang.
\newblock Few-shot class-incremental learning via class-aware bilateral distillation.
\newblock In {\em Proceedings of the IEEE/CVF Conference on Computer Vision and Pattern Recognition}, pages 11838--11847, 2023.

\bibitem{zheng2025recdreamer}
Chenxi Zheng, Yihong Lin, Bangzhen Liu, Xuemiao Xu, Yongwei Nie, and Shengfeng He.
\newblock Recdreamer: Consistent text-to-3d generation via uniform score distillation.
\newblock In {\em International Conference on Learning Representations}.

\bibitem{zhou2022forward}
Da-Wei Zhou, Fu-Yun Wang, Han-Jia Ye, Liang Ma, Shiliang Pu, and De-Chuan Zhan.
\newblock Forward compatible few-shot class-incremental learning.
\newblock In {\em Proceedings of the IEEE/CVF Conference on Computer Vision and Pattern Recognition}, pages 9046--9056, 2022.

\bibitem{zhou2022few}
Da-Wei Zhou, Han-Jia Ye, Liang Ma, Di Xie, Shiliang Pu, and De-Chuan Zhan.
\newblock Few-shot class-incremental learning by sampling multi-phase tasks.
\newblock {\em IEEE Transactions on Pattern Analysis and Machine Intelligence}, 45(11):12816--12831, 2022.

\bibitem{zhou2023uni3d}
Junsheng Zhou, Jinsheng Wang, Baorui Ma, Yu-Shen Liu, Tiejun Huang, and Xinlong Wang.
\newblock Uni3d: Exploring unified 3d representation at scale.
\newblock In {\em International Conference on Learning Representations}, pages 1--14, 2024.

\bibitem{zhu2021self}
Kai Zhu, Yang Cao, Wei Zhai, Jie Cheng, and Zheng-Jun Zha.
\newblock Self-promoted prototype refinement for few-shot class-incremental learning.
\newblock In {\em Proceedings of the IEEE/CVF Conference on Computer Vision and Pattern Recognition}, pages 6801--6810, 2021.

\bibitem{zhu2023pointclip}
Xiangyang Zhu, Renrui Zhang, Bowei He, Ziyu Guo, Ziyao Zeng, Zipeng Qin, Shanghang Zhang, and Peng Gao.
\newblock Pointclip v2: Prompting clip and gpt for powerful 3d open-world learning.
\newblock In {\em Proceedings of the IEEE/CVF International Conference on Computer Vision}, pages 2639--2650, 2023.

\end{thebibliography}
